\documentclass[10pt,letterpaper]{article}

\usepackage{cogsci}
\usepackage{graphicx}
\usepackage{natbib}
\usepackage{enumitem}
\usepackage{booktabs}
\usepackage{tabularx}
\usepackage{float}

\cogscifinalcopy

\title{The Counterexample Game:\\Iterated Conceptual Analysis and Repair in Language Models}

\author{
Daniel Drucker  \hspace{0.4em} \quad \quad \quad \quad Kyle Mahowald \\
Department of Philosophy \quad \quad Department of Linguistics \\
The University of Texas at Austin \\
\texttt{\{drucker,kyle\}@utexas.edu}
}

\begin{document}

\maketitle

\begin{abstract}
\noindent
Conceptual analysis---proposing definitions and refining them through counterexamples---is central to philosophical methodology.
We study whether language models can perform this task through iterated analysis and repair chains: one model instance generates counterexamples to a proposed definition, another repairs the definition, and the process repeats.
Across 20 concepts and thousands of counterexample-repair cycles, we find that, although many LM-generated counterexamples are judged invalid by both expert humans and an LM judge, the LM judge accepts roughly twice as many as humans do.
Nonetheless, per-item validity judgments are moderately consistent across humans and between humans and the LM.
We further find that extended iteration produces increasingly verbose definitions without improving accuracy.
We also see that some concepts resist stable definitions in general. These findings suggest that while LMs can engage in philosophical reasoning, the counterexample-repair loop hits diminishing returns quickly and could be a fruitful test case for evaluating whether LMs can sustain high-level iterated philosophical reasoning.


\end{abstract}

\noindent

\section{Introduction}
\label{sec:introduction}

 Formulating and refining conceptual analyses by subjecting them to counterexamples has long been a core part of philosophy. In book one of Plato's \emph{Republic} (c. 375), to Cephalus' proposed analysis of \textsc{justice} as paying back debts, Socrates presents this case: what if, sane, a person lends you their weapon, but crazed, they ask for it back? This method of counterexample and repair has been a core philosophical competence since at least then.

   \begin{figure}[!t]
\centering
\includegraphics[width=.93\columnwidth]{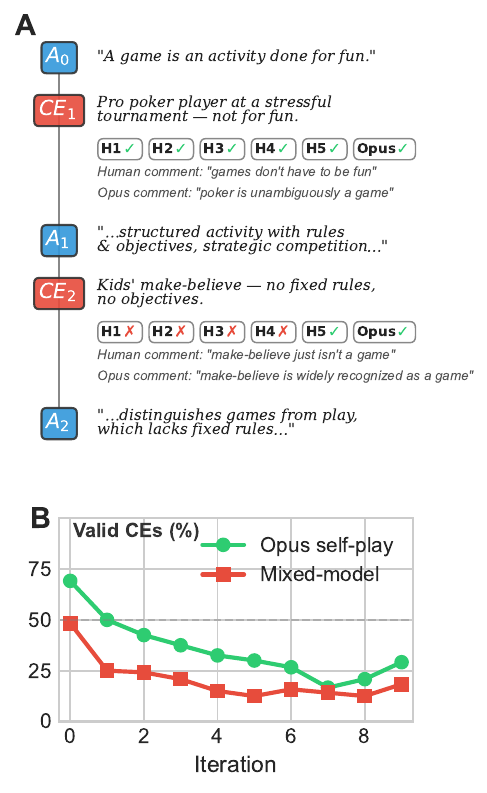}
\caption{\textbf{(A)} The counterexample-repair loop, illustrated with two iterations of a \textsc{game} chain. A seed analysis $A_0$ is challenged by $CE_1$; all five humans (H1--H5) and Opus accept it as valid. The repair $A_1$ is challenged by $CE_2$; four of the five humans reject this CE while Opus and one human accept it. Quoted snippets demonstrate reasoning. \textbf{(B)} Counterexample validity over iterations for Opus self-play vs.\ mixed-model chains, rated by Opus. Validity degrades over iterations.}
\label{fig:pipeline}
\end{figure}

This method is so well-established that graduate students in philosophy sometimes pass the time playing The Counterexample Game: Person A comes up with an analysis for a concept, Person B tries to poke holes in it, Person A tries to repair it.

Recent work with modern language models (LMs) has shown the importance of a kind of version of The Counterexample Game, in the form of adversarial ``actor-critic'' models for improving LM performance in a variety of domains \citep{shinn2023reflexion,madaan2023self}.
 In these paradigms, a model makes a proposal, and then a critic model improves it.
Likely because of the much-studied generator-discriminator gap \citep[][]{konda1999actor,goodfellow2014adversarial}, this approach has been quite successful since it allows a critic to discriminate, which then improves the generator.

Here, we bring the actor-critic model back to bear on the philosophical setting. Specifically, we test the philosophical capacities of LMs by having them play a version of The Counterexample Game. We start with a simple conceptual analysis for a concept, such as \textsc{friend}; ask another model instance (either of the same model or a different one) to give a counterexample; then ask another instance to repair the definition given that counterexample, iterating this process.

The result is a chain of definitions, counterexamples, and repairs, so the work can be thought of as an instance of iterated communication \citep{kirby_emergence_2002,kirby2014iterated,frank2012predicting}. That literature has studied the emergence of efficient linguistic behavior over iterations; here, we study whether a similar approach allows for the emergence of good conceptual analyses (relative to one's purposes), above and beyond merely increasing complexity.

Thus, our goals are threefold. First, insofar as conceptual analysis is core to philosophical thinking, we explore whether LMs can identify cogent counterexamples and generate good conceptual analyses through this iterated process. In benchmarking LMs on conceptual analyses, we join a body of work exploring conceptual analyses in LMs \citep[e.g.,][]{xu2024tip} (although none that we reviewed focused on iterated counterexample repairs). While there has been substantial work on getting LMs to think like philosophers in the ethical domain \citep{jin2022make,jin2024language,jiang2025investigating}, there has been relatively little on getting them to think like philosophers of language. Second, insofar as this is an iterated communication task, we explore the properties of the chains that emerge for different concepts, in hopes of understanding the long-range behavior of this kind of iterated repair---as well as what it can tell us about particular concepts. Third, we hope to raise more theoretical questions that would be of interest to researchers on LM and human cognition. Specifically, it could be that in general LM counterexamples don't improve with iteration. This is in some ways surprising, because human philosophers do produce better analyses and counterexamples over long periods of philosophical work.  \citep[See, e.g., the literature on causation, e.g.,][]{sep-causation-metaphysics}. Why should this process succeed when human philosophers pursue it over long periods, but not when LMs do over shorter ones?
Answering that more fully would require empirical investigations of human philosophers' ability to improve their analyses and counterexamples over shorter periods (which was outside the scope of this project).

We do, though, make progress on several fronts: We find that, while in some cases human annotators disagree as to which counterexamples are valid, there is sufficient overlap to conclude there is real signal as to which counterexamples are good or not.
We find that giving good counterexamples and repairing conceptual analyses is hard even for the best LM we tested (Claude Opus 4.5): 5 expert human annotators and Opus 4.5 both judged many model-generated counterexamples invalid.
Repeated iterations lead to looping sub-concepts (parts of the analyses given) and longer, less natural conceptual analyses. This task could, we think, serve as a challenge for developing systems that can play The Counterexample Game well and do philosophy more like philosophers.

Finally, we note that not every reader will think that there is a once-and-for-all accurate analysis for any of these concepts, especially as they are presented out of context. While one of us is more sanguine than the other about such a picture of conceptual analysis, all we rely on is {approximation} to reasonable ranges of variation in extension, upon which there is significant human convergence.

\section{Methods}

\makeatletter
\renewcommand\paragraph{\@startsection{paragraph}{4}{\z@}%
  {0.6ex \@plus 0.2ex \@minus 0.1ex}%
  {-0.6em}%
  {\normalfont\normalsize\bfseries}}
\makeatother

We study the iterative refinement of conceptual analyses through a counterexample-repair loop. A \textit{conceptual analysis} proposes necessary and sufficient conditions for a concept (e.g., ``To lie is to say something when you don't believe it's true''). A chain begins with a simple human-authored seed analysis $A_0$ (written by one of the authors, D.D., an academic philosopher), then at each step $i$ a language model generates a counterexample $CE_i$---a concrete scenario the current analysis would misclassify---and another instance revises the analysis to produce $A_i$ (Figure~\ref{fig:pipeline}).

\paragraph{Concepts} We selected 20 concepts, 10 nouns and 10 verbs (see Figure \ref{fig:ce-by-concept} for list), varying in abstractness and social complexity. Each was seeded with a simple human-authored analysis (e.g., \textsc{friend}: ``A person you like who also likes you'').

\paragraph{Models}
For the mixed-model experiments, we used three language models from different developers, randomly alternating which model performs each step: Claude 3.5 Sonnet (Anthropic), GPT-5 (OpenAI), and Gemini 2.0 Flash (Google). These were chosen to span major model families rather than to match capability tier; the goal of the mixed-model condition is to stress-test the chain dynamics under provider heterogeneity, not to benchmark a single state-of-the-art system, which we do separately in the Opus self-play setting.
We also ran {self-play experiments} using Claude Opus 4.5 (Anthropic's most capable model at time of writing), where the same model generates both counterexamples and repairs.
For Opus self-play (which is significantly more expensive), we ran 10 iterations per concept across all 20 concepts, with 6 independent chains per concept (3 with-history, 3 memoryless), totaling 1,200 counterexample-repair cycles.

\paragraph{Experimental Conditions} Each chain was run in one of two conditions: \textbf{memoryless} (each step sees only the current analysis) or \textbf{with-history} (each step sees the full prior chain of counterexamples and repairs). For the mixed-model experiments, we ran 50 iterations per concept with 3 independent chains per condition per concept (120 chains, 6,000 cycles). For counterexample generation, models received the concept name and current analysis and were asked to produce a concrete, realistic scenario that the analysis would misclassify. For repair, models received the counterexample and were instructed to revise the analysis without switching to a different concept, trivializing the analysis, or merely listing exceptions.

\paragraph{Sub-concept Tagging} To understand how \textit{conceptual content} evolves, we developed a sub-concept tagging system. For each target concept, we collected all definitions from the chain and prompted Claude 3.5 Sonnet to identify all distinct conceptual criteria appearing across them, producing 12--16 atomic, testable sub-concepts per concept (e.g., for \textsc{to lie}: \texttt{speaker\_believes\_false}, \texttt{intent\_to\_deceive}, \texttt{assertion\_context}). For each definition in the chain, we prompted Claude to determine which sub-concepts were present (explicitly or implicitly required), producing a binary presence/absence matrix.

\begin{table}[t]
\centering
\small
\setlength{\tabcolsep}{8pt}
\renewcommand{\arraystretch}{1.1}
\begin{tabular}{lccc}
\toprule
Pair & Agreement & $\kappa_{\text{val}}$ & $r_{\text{imp}}$ \\
\midrule
\multicolumn{4}{l}{\textit{Human consensus vs.\ model}}\\
Hum.\,cons.\,--\,Opus & 68\% & .42 & .59 \\
\multicolumn{4}{l}{\textit{Individual human vs.\ model}}\\
\addlinespace[2pt]
H1\,--\,Opus & 57\% & .23 & .39 \\
H2\,--\,Opus & 68\% & .39 & .53 \\
H3\,--\,Opus & 58\% & .25 & .45 \\
H4\,--\,Opus & 65\% & .33 & .37 \\
H5\,--\,Opus & 72\% & .46 & .48 \\
\multicolumn{4}{l}{\textit{Human vs.\ human}}\\
\addlinespace[2pt]
H1\,--\,H2 & 68\% & .30 & .43 \\
H1\,--\,H3 & 78\% & .34 & .51 \\
H1\,--\,H4 & 65\% & .19 & .49 \\
H1\,--\,H5 & 68\% & .26 & .52 \\
H2\,--\,H3 & 63\% & .19 & .33 \\
H2\,--\,H4 & 57\% & .10 & .23 \\
H2\,--\,H5 & 70\% & .38 & .52 \\
H3\,--\,H4 & 67\% & .23 & .41 \\
H3\,--\,H5 & 70\% & .31 & .47 \\
H4\,--\,H5 & 70\% & .36 & .29 \\
\bottomrule
\end{tabular}
\caption{Inter-rater agreement on counterexample validity (coarse yes-CE vs.\ no-CE; Cohen's $\kappa_{\text{val}}$) and Pearson correlation between raters' 1--5 importance scores ($r_{\text{imp}}$). \emph{Human consensus} is the majority vote across raters. All comparisons restricted to the same 60 human-rated items.}
\label{tab:agreement}
\end{table}

\begin{figure*}[t]
\centering
\includegraphics[width=\textwidth]{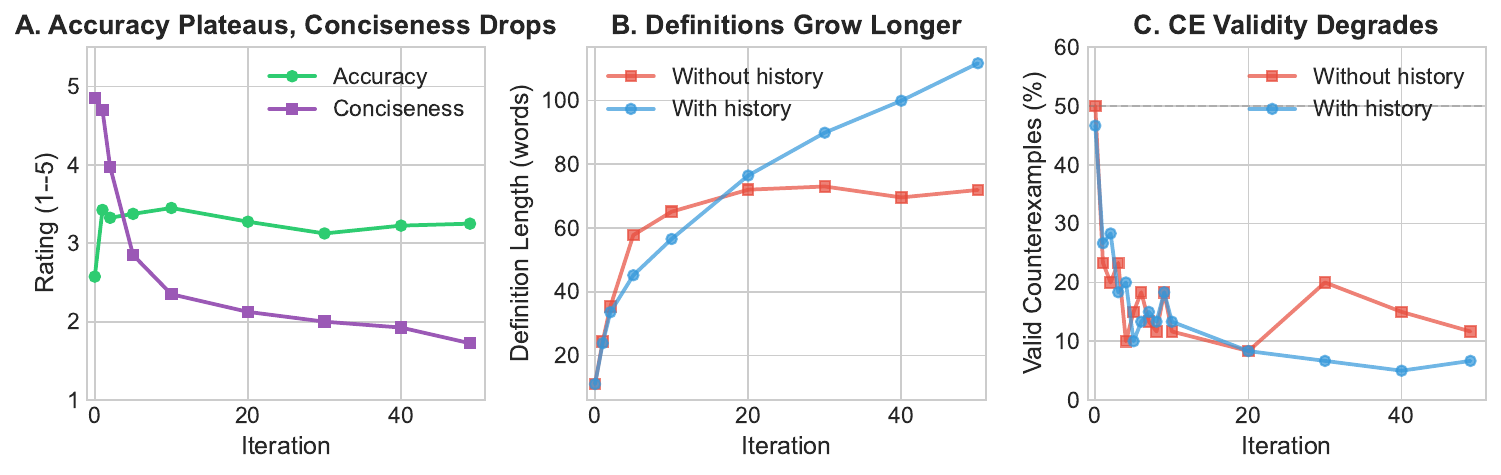}
\caption{Extended iteration over 50 rounds for our mixed-model setting. \textbf{(A)} Definition accuracy plateaus after iteration 2 while conciseness steadily degrades. \textbf{(B)} Definitions grow 9$\times$ longer. \textbf{(C)} Counterexample validity drops from 48\% to 9\%.}
\label{fig:50iter}
\end{figure*}
\paragraph{Counterexample Evaluation} We evaluate counterexample quality using both automated and human judgments (for a subset), with both using the same 4-way classification scheme. For each counterexample-analysis pair, evaluators classify the counterexample as: \textbf{valid (false positive)}, the scenario is \emph{not} an instance of the concept but the analysis would incorrectly include it (definition too broad); \textbf{valid (false negative)}, the scenario \emph{is} an instance but the analysis would incorrectly exclude it (too narrow); \textbf{invalid (handled)}, the analysis correctly classifies the scenario; or \textbf{invalid (unclear)}, the scenario is confusing, unrealistic, or doesn't engage with the analysis. For automated evaluation, we used Claude Opus 4.5 to evaluate counterexamples from mixed-model chains (1,800 CEs sampled at 15 positions across iterations 0--49) and Opus self-play chains (1,200 CEs, all 10 iterations; 6 chains per concept).

\paragraph{Human Evaluation} For human evaluation, raters (both authors and 3 philosophy graduate students from UT Austin) evaluated counterexamples sampled from iterations 1, 3, and 5 of the memoryless Opus self-play chain.
Positions were shuffled and each rater saw only ``Item X of N'' without knowing the iteration number. In addition to validity, the rater provided an importance score (1--5, from ``trivial edge case'' to ``reveals fundamental flaw'') and optional explanations or alternative counterexamples.

\paragraph{Analysis Evaluation} To track how definition quality evolves, we also evaluated analyses themselves at sampled positions (0, 1, 2, 5, 10, 20, 30, 40, 49) using Claude Opus 4.5, rating each on two dimensions (1--5 scale): \textbf{Accuracy} (does the analysis correctly capture the concept?) and \textbf{Conciseness} (is the analysis appropriately brief and non-redundant?).

\section{Results}

\subsection{Validating the LM Judge with Human Judges}

We first analyze the relationship between human judgments of the counterexamples and the ones from Claude used in the rest of the analysis.
Our primary aim in doing so is to assess whether there is meaningful signal in the quality of the counterexamples by examining the consistency between human raters and across human raters and the automated rater.

Across the 60 counterexamples (sampled from iterations 1, 3, and 5 of the Opus self-play chains, where chain-wide validity is naturally higher than later iterations; cf.\ Figure~\ref{fig:pipeline}B), the five human raters together judged $32\%$ of them valid (averaged across raters), compared to a substantially higher $60\%$ rated valid by Opus 4.5 on the same set. Because this subset is concentrated in the early, higher-validity portion of the chains, the $60\%$ Opus rate here is not directly comparable to the $36\%$ (Opus self-play) and $21\%$ (mixed) chain-wide validity rates reported below.
Of the $108$ rater--CE pairs (summed across all five raters; $5\times 60=300$ pairs total) where the human rater and Opus disagreed on validity, $96$ were cases where the human rejected the CE while Opus accepted it, against only $12$ in the reverse direction.

While the model and humans had different baseline rates of acceptance, we found moderate but substantive agreement, both among human raters and between human raters and Opus, as to which counterexamples were good.
See Table~\ref{tab:agreement}, where the \emph{human consensus} verdict for an item is the majority vote across raters on items rated by at least two humans (ties excluded). The human-consensus agreement with Opus is significantly above chance ($\kappa_{\text{val}}=0.42$; $95\%$ bootstrap CI $[+0.25,+0.59]$ over 2000 resamples).
Human-Opus agreement was clearly within the ballpark of Human-Human agreement, which itself is only in the fair-to-moderate range ($\kappa_{\text{val}}$ from $.10$ to $.38$ across the ten human pairs). This underscores that this is a task where reasonable people (and a reasonable model) can disagree, but the existence of agreement above chance points to real signal in CE quality.
A similar story emerges in the correlation between the importance scores assigned to each counterexample: humans found the same counterexamples important as did one another and as did Opus.
The correlation between the human consensus importance scores and the Opus one was moderately high ($r = .59$).

Because Opus accepts roughly twice as many counterexamples as the average human, the absolute LM-judge validity rates we report below are probably best read as upper bounds on what humans would call valid.

In some cases, it was clear where counterexamples were strong. An analysis of \textsc{game} said ``A game is an activity done for fun.'' All humans and Opus rated as valid a counterexample pointing out the case of a stressed-out professional poker player who is not enjoying playing poker but does it anyway. They all agreed poker was still a game even if not an activity done for fun by a particular person.
(Note that we called the central philosophical exercise in this paper ``The Counterexample Game'', but leave it to the reader whether it is fun.)

In other cases, all parties agreed that some counterexamples were flawed. The analysis for \textsc{neighbor} suggested ``close proximity'' was required to be a neighbor. The counterexample gave an example of a rural family where the nearest ``neighbor'' is 2 miles away even though 2 miles is not close proximity.
The human raters and Opus rejected this counterexample, arguing that close proximity is relative and 2 miles can well be considered close.

In other cases, most humans \textit{disagreed}
with Opus. For instance, regarding a counterexample for \textsc{game} involving kids' make-believe with no fixed rules, four of the five humans said that rule-less make-believe is not a game; the fifth conceded that ``it is plausible to say that they are.'' Opus took it as a valid counterexample, stating in its explanation: ``Make-believe/imaginative play is widely recognized as a type of game (developmental psychologists, game theorists, and ordinary usage all categorize it as `playing a game').''

Reading the human comments alongside Opus's reasoning points to some systematic differences. Opus almost always quotes specific clauses of the analysis and proposes refinements (``the analysis fails to capture the epistemic conditions necessary
  for genuine consent''; ``too narrow in requiring orientation
  toward internal purposes''), and sometimes invokes external
  authority---as with the psychologists and game theorists in the \textsc{game} make-believe case above.
  Humans, by contrast, more often hand down a brief intuitive
  verdict: ``classic!'', ``seems totally fine'', and ``lol cleverly
  includes bread bowl'' (in response to a CE proposing a bread bowl as a counterexample to the analysis of \textsc{sandwich}).
  Humans also hedge more (``edge case'', ``I'm not
  entirely sure'', ``I lean towards it being a good CE but
  weakly'').
  Some of this is that the LM is more verbose and formal, but the pattern also points to deeper differences in how humans and the LM approach the task.

To be sure, there were also many cases where humans disagreed with one another, highlighting some genuine points of disagreement as to what constitutes a good counterexample---philosophers don't always agree on definitions.
An analysis of \textsc{accuse} involved ``to say someone did something wrong'', and the counterexample involved a mother telling a child ``You left the milk out again.''
3 of the 5 human annotators treated the case as a valid counterexample; 1 considered it an accusation correctly captured by the analysis; and the remaining rater flagged the analysis itself as ambiguous, distinguishing between ``to say that someone did something which is, in fact, wrong'' vs. ``to say, as it were, `someone did something and that thing is wrong.'''
Identifying and resolving these kinds of disagreements are central to the exercise.
But, for our purposes, we judge there to be sufficient signal in the task, and enough agreement with the LM judge, to proceed with further automated analysis.

\subsection{Quality of Analyses and Counterexamples}

We now quantitatively and qualitatively evaluate the success of the counterexample-repair process across iterations, using the LM judge. We start with analysis quality over time, for the 50-iteration mixed-model case both with and without history, as rated by Opus 4.5 (see Figure~\ref{fig:50iter}). First, the accuracy of the analysis (as rated) rises modestly in the first 2 iterations and then plateaus, while conciseness (as rated) drops steadily from 4.9 to 1.7 (Panel A). Second, definitions grow substantially longer, from 11 words initially to 95 words on average (Panel B). Third, counterexample validity drops precipitously---from 48\% at iteration 0 to just 9\% by iteration 49 (Panel C). The widening gap between accuracy and conciseness suggests that models are producing increasingly verbose definitions without improving accuracy. Qualitatively, the longer definitions tend to become increasingly list-like: handling special cases but without necessarily making a better and more stable definition.
History and no-history conditions show similar patterns, suggesting that access to prior iterations does not help.

One possible reason for the definitions not improving could be low counterexample quality.
If models generate spurious counterexamples, repairs may be responding to phantom problems.
We find low overall counterexample validity, and substantial differences in validity between conditions.
Namely, Opus self-play produces higher rates of valid counterexamples compared to mixed-model chains (36\% vs 21\%), likely because it is a more capable model, although also possibly due to negative effects of heterogeneity in mixed chains.

Counterexample validity degrades substantially over the course of a chain (Figure~\ref{fig:pipeline}B): 69\% of CEs are valid at position 0 (the first counterexample to the seed), 42--50\% at positions 1--2, and just 17--30\% by positions 5--9. This pattern suggests that early counterexamples successfully identify genuine flaws in simple seed analyses.
But, as definitions become more refined, models struggle to find valid counterexamples.

Providing models with the full history of prior counterexamples and repairs does not seem to improve counterexample validity: Opus self-play produces nearly identical validity rates in history and memoryless conditions (35.0\% vs.\ 36.0\%), as do the mixed-model chains  (16.7\% vs.\ 18.0\%).

\subsection{Concept-Level Variation}

\begin{figure}[t]
\centering
\includegraphics[width=1\columnwidth]{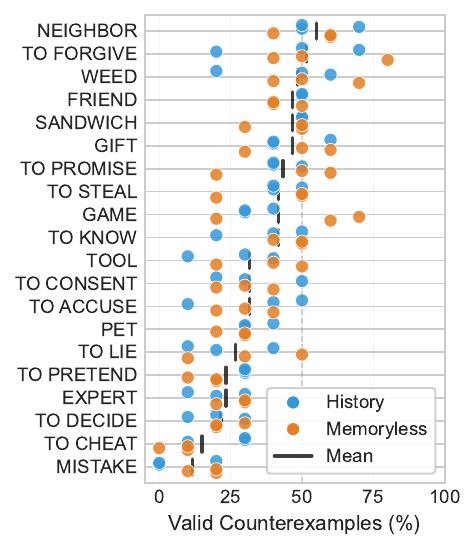}
\caption{Counterexample validity rate per concept across all 6 Opus self-play chains (3 with history, 3 memoryless). Each dot represents one chain's validity rate; black lines show the mean. Concepts are sorted by mean validity. Concept difficulty is consistent across chains.}
\label{fig:ce-by-concept}
\end{figure}

We next turned to exploring concept-level variation, to understand differences across concepts in the ease of developing good analyses and the ease of finding valid counterexamples.
We observe that counterexample validity varies substantially across concepts (Figure~\ref{fig:ce-by-concept}). Some concepts appear easier to attack, while others resist valid counterexamples. Among the highest validity counterexamples (easiest to attack) were:
\textsc{neighbor} (55\%), \textsc{to forgive} (52\%), \textsc{weed} (48\%), \textsc{friend} (47\%).
Among the lowest validity counterexamples (hardest to attack) were:
\textsc{mistake} (12\%), \textsc{to cheat} (15\%), \textsc{to decide} (22\%), \textsc{expert} (23\%).

 Figure~\ref{fig:ce-by-concept} shows that these concept-level differences are reasonably consistent across independent chains: concepts that are easy (or hard) to attack in one chain tend to be easy (or hard) in all chains, regardless of whether the model has access to prior history. This suggests that concept difficulty reflects genuine properties of the concepts rather than noise.

\begin{figure*}[t]
\centering
\includegraphics[width=.95\textwidth]{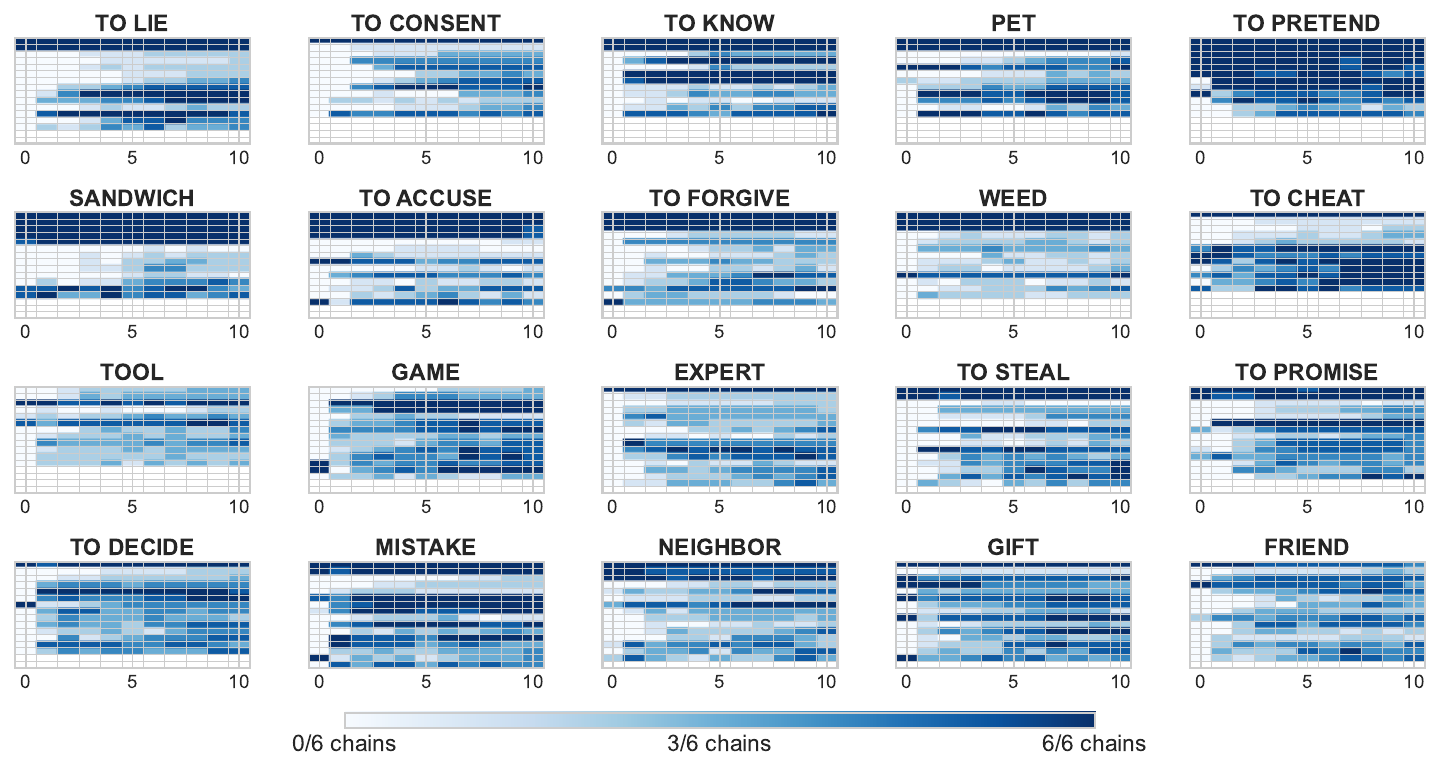}
\caption{Sub-concept presence across iterations for all 20 concepts, aggregated over 6 independent chains. Color intensity indicates the fraction of chains where a sub-concept is present (white = 0/6, dark blue = 6/6).
Each row is one sub-concept; columns show iterations 0--10. Within each concept, all sub-concepts are shown; concepts with fewer sub-concepts have blank rows below.}
\label{fig:subconcept-evolution}
\end{figure*}

To explore where models go wrong and why counterexamples and repairs fail to enter a virtuous loop of self-improvement, we examine how the \textit{content} of definitions evolves across iterations using our sub-concept tagging methodology (see Table~\ref{tab:subconcepts} for a sample of sub-concepts). We focus this analysis on the Opus-only chains.

Concepts differ in how stable their definitions remain across iterations (Figure~\ref{fig:subconcept-evolution}).
We examine oscillation: how often the same concept changes state over a chain.
Consider \textsc{expert}.
Its sub-concept \texttt{principle comprehension} (whether the expert requires understanding principles as opposed to just having knowledge) comes and goes across iterations, winking in and out around 4 times per chain.
On the other hand, the sub-concept \texttt{relinquish resentment} is present in almost every chain for \textsc{to forgive}.

Figure \ref{fig:subconcept-evolution} shows this pattern at a macro-scale: some concepts have highly stable sub-concepts, and others don't.
Taken together with Figure \ref{fig:ce-by-concept}, we observe that concepts pattern differently.
Why? First, {genuine difficulty}: it is likely some concepts are inherently harder to define, leading  to both definitional instability and difficulty generating valid counterexamples. Second, {moving target}: when definitions shift dramatically between iterations, counterexamples generated against one version may not apply to the next. Third, {prevalence in pre-training}: \textsc{to lie}, \textsc{to consent}, and \textsc{to know} have received a lot of attention in the philosophical literature, so the ability to formulate pretty good conceptual analyses of these may simply fall out of that fact.

\begin{table}[t]
\footnotesize
\centering
\begin{tabularx}{\columnwidth}{@{}lX@{}}
\toprule
\textbf{Word} & \textbf{Sub-concepts}  \\
\midrule
\textsc{to accuse} & attributes responsibility, identifies wrongdoing, demands accountability, direct assertion \\
\textsc{expert} & deep domain knowledge, pattern recognition, principle comprehension, contextual application\\
\textsc{to cheat} & intentional violation, rule-governed activity, unfair advantage, prohibited means, competitive context \\
\textsc{friend} & mutual affection, shared history, survives separation, mutual recognition, enduring commitment, goodwill \\
\bottomrule
\end{tabularx}
\caption{Selected sub-concepts for a sample of concepts.}
\label{tab:subconcepts}
\end{table}

\section{Discussion}

For generations, philosophers have proposed and refined concepts iteratively---often leading to better conceptual analyses.
In our Counterexample Game, we do not see similar long-term improvement over iterations, only longer definitions.
Of course, an apples-to-apples comparison would require asking humans to generate counterexamples using the same paradigm.
But we find LMs' failures suggestive that they do not benefit from this kind of iterative process.

As such, our findings also bear on the growing literature on LM self-improvement through actor-critic loops \citep{shinn2023reflexion,madaan2023self}. Such approaches assume that the critic can reliably identify errors. In our setting, the critic (counterexample generator) often fails, and when it does produce valid counterexamples, the generator (repair model) often responds by adding a laundry list of exceptions, rather than achieving genuine conceptual insight. This parallels findings in iterated learning, where some tasks benefit from extended iteration while others plateau or degrade \citep{kirby2014iterated}.

One striking finding is that LMs' analyses get longer but not better.
A possibility is that LMs use the length of a given formulation of an analysis (or counterexample) as a measure of quality \citep[the so-called verbosity bias;][]{zheng2023judging}.
If so, that could partially explain their performance issues. It would also suggest a way of improving LMs' performance on this task, by attempting to correct for a ``long text'' bias. It would be interesting to see if this bias affected performance on other philosophical tasks, like argument plausibility evaluation.

Our results also serve not just as a test of LMs as philosophers, but as a lens onto inter-concept differences.
Concept heterogeneity likely makes some concepts harder than others.
For instance, \textsc{game} is a paradigmatic example of a concept that some philosophers \citep[e.g.,][]{wittgenstein1953philosophical} have argued resists definition in terms of necessary and sufficient conditions, instead exhibiting ``family resemblance'' among instances \citep[but see][for a defense of definitional analysis]{Suits1978}.

\section{Conclusion}

We introduced LMs to The Counterexample Game---iterated counterexample-repair for conceptual reasoning.
While well-trained people can disagree as to what makes a good counterexample, we found significant convergence in the task for humans and models.
Extended iteration produced longer analyses, but not necessarily better ones.
We see our work as proof of concept for using philosophical methodology to study LMs, as well as for generating conceptual analyses and insights.

\section{Acknowledgments}
K.M. acknowledges support from National Science Foundation grant 2339729.
For help with philosophical annotation at different points in the project, we thank Sam Cantor, Henry Curtis, Johan Gustafsson, Nathan Hagan, Andrew Kovacs, and
Danny Shea. We thank Sybil Li for helpful discussions, particularly in the early stages of the project.
We used Claude Code for building an annotation framework, helping generate concepts, writing code, and summarizing results. Other uses (e.g., subconcept labeling, LM as judge) are described in text.

\setlength{\bibhang}{0.125in}

\bibliographystyle{apalike}
\bibliography{ce}

\end{document}